\documentclass[unnumsec,webpdf,contemporary,large,table]{oup-authoring-template}%

\graphicspath{{Fig/}}

\theoremstyle{thmstyleone}%
\theoremstyle{thmstyletwo}%
\theoremstyle{thmstylethree}%

\usepackage{tabularray}
\usepackage{booktabs}
\usepackage{pifont}
\usepackage{breakurl}

\begin{document}

\journaltitle{Journal Title Here}
\DOI{DOI HERE}
\copyrightyear{2022}
\pubyear{2019}
\access{Advance Access Publication Date: Day Month Year}
\appnotes{Paper}

\firstpage{1}


\title[pyclipse, a deidentification toolkit]{\emph{Pyclipse}, a library for deidentification of free-text clinical notes}

\author[1,$\ast$]{Callandra Moore\ORCID{0009-0008-3801-3137}}
\author[2]{Jonathan Ranisau\ORCID{0000-0002-1178-7510}}
\author[2,3]{Walter Nelson\ORCID{0000-0002-5216-7617}}
\author[2,4,5,6]{Jeremy Petch\ORCID{0000-0003-1614-1046}}
\author[7]{Alistair Johnson\ORCID{0000-0002-8735-3014}}

\authormark{Moore et al.}

\address[1]{\orgdiv{Child Health Evaluative Sciences}, \orgname{Hospital for Sick Children}, \orgaddress{\state{Ontario}, \country{Canada}}}
\address[2]{\orgdiv{Centre for Data Science and Digital Health}, \orgname{Hamilton Health Sciences}, \orgaddress{\state{Ontario}, \country{Canada}}}
\address[3]{\orgdiv{Department of Statistical Sciences}, \orgname{University of Toronto}, \orgaddress{\state{Ontario}, \country{Canada}}}
\address[4]{\orgdiv{Institute of Health Policy, Management and Evaluation}, \orgname{University of Toronto}, \orgaddress{\state{Ontario}, \country{Canada}}}
\address[5]{\orgdiv{Division of Cardiology, Department of Medicine}, \orgname{McMaster University}, \orgaddress{\state{Ontario}, \country{Canada}}}
\address[6]{\orgdiv{Population Health Research Institute}, \orgname{McMaster University}, \orgaddress{\state{Ontario}, \country{Canada}}}
\address[7]{\orgdiv{Dalla Lana School of Public Health}, \orgname{University of Toronto}, \orgaddress{\state{Ontario}, \country{Canada}}}

\corresp[$\ast$]{Corresponding author. \href{email:callandra.moore@sickkids.ca}{callandra.moore@sickkids.ca}}

\received{Date}{0}{Year}
\revised{Date}{0}{Year}
\accepted{Date}{0}{Year}

\abstract{Automated deidentification of clinical text data is crucial due to the high cost of manual deidentification, which has been a barrier to sharing clinical text and the advancement of clinical natural language processing. However, creating effective automated deidentification tools faces several challenges, including issues in reproducibility due to differences in text processing, evaluation methods, and a lack of consistency across clinical domains and institutions.
To address these challenges, we propose the \emph{pyclipse} framework, a unified and configurable evaluation procedure to streamline the comparison of deidentification algorithms. \emph{Pyclipse} serves as a single interface for running open-source deidentification algorithms on local clinical data, allowing for context-specific evaluation.
To demonstrate the utility of \emph{pyclipse}, we compare six deidentification algorithms across four public and two private clinical text datasets. We find that algorithm performance consistently falls short of the results reported in the original papers, even when evaluated on the same benchmark dataset. These discrepancies highlight the complexity of accurately assessing and comparing deidentification algorithms, emphasizing the need for a reproducible, adjustable, and extensible framework like \emph{pyclipse}.
Our framework lays the foundation for a unified approach to evaluate and improve deidentification tools, ultimately enhancing patient protection in clinical natural language processing.}

\maketitle

\section{Introduction}
The advent of large, open access text corpora and the empirical efficacy of neural networks have driven advances in model performance in natural language processing. In healthcare, the increasing volume of electronic health record (EHR) data from hospitals provides a rich source of information on patient state, treatment, and outcome in the form of semi-structured and unstructured text \cite{norgeot2020, mayer2009}. Barriers to sharing clinical text, however, have stifled the application of natural language processing in the medical domain. In particular, clinical text from EHRs cannot be fully leveraged for scientific research without first minimizing the risk of harm to patients \cite{ahmed2021}. 

To use identifiable data for analysis, researchers typically must obtain the informed consent of individual patients, the approval of an internal review board, or both. In many jurisdictions these requirements can be waived when the data are sufficiently de-risked through the process of deidentification, where patient identifiers are substituted with surrogates or removed entirely. In the United States, the Health Insurance Portability and Accountability Act (HIPAA) provides federal protection for protected health information (PHI). HIPAA permits sharing of non-individually identifiable data, determined either through “Expert Determination” or via a “Safe Harbor” provision which outlines the eighteen specific identifiers that must be removed to consider a dataset “deidentified”. These identifiers include patient names; medical record numbers; dates; ages over 89 years; any contact information; geographic locations; and any other identifier, code, or characteristic \cite{hipaa}. Deidentified datasets are safer to share among multiple research groups which simplifies complementary analysis, comparisons, and replication \cite{henriksson2014, datta2020, ahmed2021}.

There have been a few shared tasks evaluating EHR deidentification methods, in particular those hosted in 2006 \cite{uzuner2007} and 2014 \cite{stubbs2015automated} by the Informatics for Integrating Biology and the Bedside (I2B2) project and in 2016 by the Centers of Excellence in Genomic Science Neuropsychiatric Genome-Scale and RDoC Individualized Domains (N-GRID) \cite{stubbs2017}. 

These public datasets have been extremely beneficial in providing a consistent evaluation set for community comparison.
Despite the utility of these benchmark datasets, limitations remain. Performance of existing algorithms remains difficult to reproduce even on the established benchmark. For instance, Norgeot et al. (2020), despite using the i2b2 2014 dataset as the basis and benchmark for their PHIlter deidentification algorithm, first preprocess the provided gold-standard annotations to remove entities not required by HIPAA's Safe Harbor provision \cite{norgeot2020}, complicating reproduction of their results (see \url{https://github.com/BCHSI/philter-ucsf/issues/11}).

Although privately held deidentified clinical corpora range in the millions of notes, the largest benchmark dataset contains less than 2,000 notes. These datasets contain only synthetic PHI, as publicly releasing real PHI would be a legal violation. Generation of synthetic PHI is a challenging endeavour on its own, and even with great care will result in a distribution shift. Finally, current benchmark datasets cover a subset of clinical note domains (discharge summaries, nursing notes, pathology reports), and omit large swaths of other note types (primary care notes, radiology reports, consult notes, specialist notes, and so on). These other note types may challenge generalizability of proposed models.

We identified the need for a consistent framework which solved the above limitations.
Our proposed framework simplifies the comparison of performance of deidentification algorithms on a single dataset with a consistent evaluation procedure and solves the above limitations. We provide adapters to run existing open-source deidentification algorithms and compare their performance. Most importantly, our framework allows application and comparison of these algorithms on local clinical data, allowing context-specific evaluation most relevant to the end user.

This research was reviewed and approved by the Research Ethics Board of the Hospital for Sick Children and the Hamilton Integrated Research Ethics Board. 

\section{Methods}

We created \emph{pyclipse}, a Python-based package for evaluating deidentification algorithms on public and private datasets. The goal of \emph{pyclipse} is two-fold. First, it provides a consistent format for datasets and utilities to map to this dataset. This simplifies evaluation of algorithms across the publicly available benchmark datasets, and ensures similar processing occurs for privately held datasets. Second, \emph{pyclipse} provides a structure for evaluating algorithms on these datasets, as well as wrappers for the currently available open-source algorithms. \emph{Pyclipse} is open source and available under a permissive license  \url{https://github.com/kind-lab/pyclipse}.

\emph{Pyclipse} contains a number of utilities common in the evaluation of deidentification algorithms:
\begin{itemize}
    \item Standardized dataset format using parquet.
    \item Method to map PHI entity types from different annotation styles to a common set (e.g., ``medicalrecordnumber" to ``ID"). This method can be customized by the user.
    \item A set of Python wrappers for the deidentification algorithms tested. These wrappers enable the usage of algorithms from a variety of languages (Python, Java, and Perl) with other utilities in \emph{pyclipse} and may serve as examples for the evaluation of future deidentification algorithms. 
    \item Evaluation framework which can be applied to any combination of deidentification algorithm and dataset. Evaluation can be easily configured to accomodate different tokenizers and other parameters. This includes multiple methods to merge annotations across algorithms.
    \item Mechanism to scrub PHI from documents using predicted annotations.
    \item Tool to visualize the false positives, false negatives, and true positives for a predicted set of annotations versus the gold standard.
    \item Utility to save token-level token indices and annotations independent of text to enable the deletion of all documents and PHI but retain the ability to run evaluations and ensure replicability. 
\end{itemize}

These utilities contribute to a standardized and replicable evaluation process, facilitate the comparison of different algorithms and datasets, and provide a clear pathway and API for adapting future algorithms.

\subsection{Datasets available}

We collected five datasets of free-text notes written during routine clinical practice and converted existing publicly available datasets into our standard format. Four of these datasets are publicly available after the respective data use agreement is signed. Due to restrictions on the distribution of data, we cannot redistribute reformatted data without express permission. Instead, we provide clear instructions and code for transforming datasets into our format.

\begin{itemize}
    \item i2b2 2006 Corpus: 889 deidentified discharge summaries shared as part of the 2006 i2b2 Challenge. Includes challenge annotations, training and test sets, and ground truth \cite{uzuner2007}.
    \item i2b2 2014 Corpus: 1,304 medical records for 296 patients, shared as part of the 2014 i2b2 Challenge \cite{stubbs2015annotating}.
    \item PhysioNet Corpus version 2: 2,434 nursing notes collected from patients admitted to intensive care units at the Beth Israel Deaconess Medical Center, Boston, Massachusetts, USA \cite{neamatullah2008}. Gold-standard annotations are from an improved set released by Google \cite{hartman2020customization}. 
    \item OpenDeID Corpus: 2,100 pathology reports from 1,833 cancer patients from four urban Australian hospitals \cite{jonnagaddala2021}.
    \item RR-SK: 1,083 radiology reports collected from the Hospital for Sick Children (HSC) in Toronto, Ontario, Canada. This corpus is not currently open-source.
    \item RR-HHS: 50 radiology reports collected from Hamilton Health Sciences (HHS) in Hamilton, Ontario, Canada. This corpus is not currenty open-source. 
\end{itemize}

Where specified, we used the author-defined test set for evaluation. For the RR-HHS dataset, the test set is composed of all fifty currently annotated notes. Where no test set is available (e.g., the newly collected RR-SK dataset), we create a random subsample and publish these identifiers to establish a common test set.
These datasets have all been annotated with a set of gold-standard annotations against which they can be evaluated. However, the annotation strictness differs in a few key ways between the corpi, namely whether lone years, ages less than 90, and professions are labeled. These differences are detailed in Table \ref{tab:tool_ann_types}. 
Summary characteristics of the datasets are outlined in Table \ref{tab:dset_PHI_types}. As shown in this table, the types and distributions of annotations associated with each dataset vary; however, all labels have been mapped to one of name, profession, age, id, date, location, or contact to enable comparisons across datasets. 

\begin{table*}[ht!]
\centering
\caption{Breakdown of protected health information types and subtypes in the five datasets under evaluation. Values in parentheses are the proportion of that entity type to the total PHI tokens. }
\begin{tabular}{c|cccccc}
\hline
\toprule
PHI type & i2b2 2014 & i2b2 2006 & PhysioNet v2 & OpenDeID & RR-SK & RR-HHS \\ \midrule
\rowcolor[gray]{0.9} 
name & 5559 (0.18) & 2925 (0.24) & 228 (0.35) & 8221 (0.25) & 790 (0.21) & 1143 (0.35) \\
name & - & - & - & - & 789 & 1143 \\
doctor & 3654 & 2410 & - & 5803 & - & - \\
patient & 1813 & 515 & - & 2418 & - & - \\
username & 92 & - & - & - & - & - \\
hcpname & - & - & 138 & - & - & - \\
relativeproxyname & - & - & 68 & - & - & - \\
ptname & - & - & 20 & - & - & - \\
ptnameinitial & - & - & 2 & - & - & - \\
misc & - & - & - & - & 1 & - \\ \hline
\rowcolor[gray]{0.9} 
profession & 345 (0.01) & - & - & - & 3 (0.0002) & 0 (0.00) \\ \hline
\rowcolor[gray]{0.9} 
location & 3025 (0.10) & 1921 (0.16) & 85 (0.13) & 7171 (0.22) & 130 (0.03) & 206 (0.06) \\
location & - & 242 & 85 & - & - & - \\
loc & - & - & - & - & 41 & - \\
department & - & - & - & 2704 & - & 206 \\
hospital & 1604 & 1679 & - & 866 & - & - \\
organization & 154 & - & - & 1 & - & - \\
org & - & - & - & - & 89 & - \\
street & 416 & - & - & 1944 & - & - \\
state & 206 & - & - & 496 & - & - \\
city & 347 & - & - & 640 & - & - \\
country & 130 & - & - & 2 & - & - \\
zip & 148 & - & - & 504 & - & - \\
location-other & 20 & - & - & 14 & - & - \\ \hline
\rowcolor[gray]{0.9} 
age & 813 (0.03) & 3 (0.0002) & - & 59 (0.001) & 268 (0.07) & 69 (0.02)
\\ \hline
\rowcolor[gray]{0.9} 
date & 19701 (0.62) & 5195 (0.43) & 293 (0.45) & 12778 (0.39) & 2551 (0.67) & 1175 (0.37) \\
date & 19701 & 5195 & 286 & 12778 & 2511 & 1100 \\
dateyear & - & - & 7 & - & - & - \\
time & - & - & - & - & 40 & 75 \\ \hline
\rowcolor[gray]{0.9} 
id & 1592 (0.05) & 1714 (0.14) & 2 (0.003) & 4462 (0.14) & 16 (0.004) & 704 (0.21) \\
device & 14 & - & - & - & - & - \\
idnum & 560 & - & - & 2371 & 16 & 704 \\
medicalrecord & 1018 & - & - & 2091 & - & - \\
id & - & 1714 & - & - & - & - \\
other & - & - & 2 & - & - & - \\ \hline
\rowcolor[gray]{0.9} 
contact & 662 (0.02) & 245 (0.02) & 50 (0.08) & 1 (0.0000) & 30 (0.008) & 0 (0.00) \\
email & 5 & - & - & - & - & - \\
fax & 12 & - & - & - & - & - \\
phone & 645 & 245 & 50 & 1 & - & - \\
contact & - & - & - & - & 30 & - \\ \midrule
phi total & 31697 & 12003 & 658 & 32692 & 3788 & 3297 \\
total tokens & 435859 & 184731 & 103078 & 544877 & 154491 & 17969 \\
total documents & 514 & 701 & 502 & 702 & 432 & 50 \\ \hline
\end{tabular}
\label{tab:dset_PHI_types}
\end{table*}

\subsection{Dataset format}

We adopted parquet, an open source columnar based binary format used for efficient storage of large datasets \cite{parquet}. 
In the \emph{pyclipse} format, the data is stored in three parquet datasets: a dataset of the document text, a dataset of annotations partitioned by algorithm, and an optional dataset of gazetteers. \emph{Pyclipse} contains utilities to convert data into this format from various standoff formats, including from individual text files and eXtended Markup Language (XML) formats used by the i2b2 datasets \cite{uzuner2007, stubbs2015annotating}.

Often, annotation methods require the use of gazetteers to form a customizable dictionary or block list for annotation. Though none of these datasets come with gazetteers, they were added for the i2b2 2014, i2b2 2006, and OpenDeID corpi via inspection of the corpus text and the annotation files for the patients’ first and last names and medical record number.

\subsection{Algorithms}

We implemented wrappers for five openly available deidentification algorithms: Philter-UCSF \cite{norgeot2020}, TiDE \cite{datta2020}, PyDeID \cite{johnson2019}, PhysioNet-deid \cite{neamatullah2008}, and Transformer-DeID \cite{moore2023}. These algorithms adopt various approaches to deidentification including the use of supervised machine learning (ML) with labeled training data, rule-based approaches, and dictionaries of known entities.
A comparison of these algorithms is provided in Table \ref{tab:deid_tools_meta}. Similar to the datasets described above, each deidentification algorithm differs slightly in which entities are considered protected, including the annotation of professions, ages less than 90, non-patient names, large geographic locations such as country, lone years, and organizations. These differences are highlighted in Table \ref{tab:tool_ann_types}. We harmonize entity types output by each algorithm for consistency.

\begin{table*}[ht!]
\centering
\caption{Information for each deidentification algorithm.}
\label{tab:deid_tools_meta}
\centering
\begin{tabular}{|m{2.5cm}|m{4cm}|c|m{2.5cm}|c|m{3.4cm}|}
\hline
Dataset & Repository & Citation & Method & Gazetteers Used & Source Dataset \\ \hline
Transformer-DeID\newline \hspace*{2pt} DistilBERT \newline \hspace*{2pt} RoBERTa & \url{https://github.com/kind-lab/transformer-deid/} & \cite{moore2023} & ML & No & i2b2 2014 \\
Philter & \url{https://github.com/BCHSI/philter-ucsf} & \citep{norgeot2020} & Rule-based and dictionaries & No & i2b2 2014 \\
PyDeID & \url{https://github.com/mit-lcp/pydeid} & \cite{johnson2019} & Rule-based and dictionaries & No & Beth Israel Free Text Notes \\
TiDE & \url{https://github.com/susom/tide} & \citep{datta2020} & ML, rule-based, and dictionaries & Yes & MTSamples Laparoscopic Gastric Bypass Consult (augmented) \\
PhysioNet & \url{https://www.physionet.org/content/deid/1.1/} & \cite{neamatullah2008} & Rule-based and dictionaries & No & PhysioNet \\ \hline
\end{tabular}
\end{table*}

\begin{table*}[ht!]
\caption{Discrepancies between entities considered protected by each deidentification algorithm and dataset.}\label{tab:tool_ann_types}
\scriptsize
\begin{tabular}{l|p{2.4cm}|p{1.4cm}p{1.4cm}p{2.4cm}p{3cm}p{1.3cm}p{1.8cm}}\toprule
& & profession? & ages $<$ 90? & non-patient names? & large-granularity geographic locations? & year? & organization? \\\midrule
& HIPAA Safe Harbor &  &  &  &  &  &  \\\midrule
\parbox[t]{2mm}{\multirow{6}{*}{\rotatebox[origin=c]{90}{algorithm}}} & Transformer-DeID & \ding{51} &\ding{51} &\ding{51} &\ding{51} &\ding{51} &\ding{51} \\
& Philter & & &\ding{51} & & & \\
& PyDeID & &\ding{51} &\ding{51} &\ding{51} & &\ding{51} \\
& TiDE & & &\ding{51} &\ding{51} & & \\
& PhysioNet & & &\ding{51} & &\ding{51} &\ding{51} \\ \midrule
\parbox[t]{2mm}{\multirow{6}{*}{\rotatebox[origin=c]{90}{dataset}}} & i2b2 2014 &\ding{51} &\ding{51} &\ding{51} &\ding{51} &\ding{51} &\ding{51} \\
& i2b2 2006 & & &\ding{51} & & &\ding{51} \\
& PhysioNet & & &\ding{51} & & &\ding{51} \\
& OpenDeID & &\ding{51} &\ding{51} &\ding{51} &\ding{51} &\ding{51} \\
& RR-SK &\ding{51} &\ding{51} &\ding{51} &\ding{51} &\ding{51} &\ding{51} \\
& RR-HHS &\ding{51} &\ding{51} &\ding{51} &\ding{51} &\ding{51} &\ding{51} \\
\bottomrule
\end{tabular}
\end{table*}

\subsection{Evaluation}

Our evaluation pipeline is composed of three phases. First, tokenization is applied to separate the text into individual tokens which will form the basis of later comparisons. Many named entity recognition (NER) datasets are distributed with individual words separated, e.g. CoNLL \cite{tjong2003}. As our data format stores the raw text, tokenization becomes an important component of the evaluation pipeline.
By default, tokenization is done with the WordPunctTokenizer in the Python Natural Language ToolKit (NLTK) module \cite{bird2009}. This tokenizer separates tokens on whitespace and punctuation. As evaluation is sensitive to the type of tokenizer used, we allow for the configuration and comparison of different tokenizers.
Second, an entity mapper is used to allow for various evaluation scenarios.
For example, the HIPAA-Strict evaluation scenario only evaluates performance on the exact entities defined in the HIPAA Safe Harbor provision, i.e. it does not evaluate professions, doctor names, and so on. Alternatively, binary evaluation simplifies all entities into either a PHI or non-PHI category.
Finally, measures of model performance are calculated using the final labels. These include precision, recall, and F1-scores.

\subsection{Experiments}

Using the established framework of \emph{pyclipse}, we evaluated all algorithms on the set of benchmark datasets to establish reference performance measures.
Each algorithm was applied to each dataset, and the resultant annotations were stored.
These annotations were converted to a binary evaluation from individual entities (i.e. from name, date and so on to PHI vs. non-PHI) to enhance comparability of the algorithms. Text was tokenized using the default NLTK tokenizer.

In addition to the binary evaluation, we perform a separate analysis focusing on recall of the ``name'' entity.
PHI covered by HIPAA’s Privacy Rule differ in severity, with the most severe being those entities which would allow for reidentification of patients absent any organizational access, such as name \cite{mayer2009}.
Further, this entity type is the most consistently annotated across all corpi and models.

Finally, we compare the time taken to run each algorithm on fixed hardware using CPU based inference.

\section{Results}
Table \ref{tab:f1} shows the F1-score of the six deidentification algorithms over the six datasets examined, evaluated token-wise after binarizing labels to PHI or non-PHI. In this case, a predicted label for a protected entity is considered correct if the model identifies it as PHI, regardless of specific PHI subtype identified by the algorithm. 

No model-dataset combination achieved an F1-score greater than 0.95. Notably, none of the algorithms outperformed the state-of-the-art F1-score of 0.977 achieved during the i2b2 2014 deidentification challenge \cite{stubbs2015automated}. According to the F1-score, the Transformer-DeID RoBERTa model is the best performing algorithm for all datasets with the exception of the RR-SK corpus, on which it was slightly outperformed by TiDE, PhysioNet, and DistilBERT. The best F1 performance (0.924) was achieved by the Transformer-DeID RoBERTa model on the i2b2 2014 dataset, which is complementary to its training dataset. 

\begin{table*}[ht]
\centering
\caption{F1-scores for each deidentification algorithm over each dataset for all PHI types. Performance is calculated token-wise using binary PHI or non-PHI labels.}
\label{tab:f1}
\begin{tabular}{lrrrrrr}
\toprule
 & i2b2 2014 & i2b2 2006 & PhysioNet v2 & OpenDeID & RR-SK & RR-HHS \\
\midrule
DistilBERT & 0.905 & 0.710 & 0.461 & 0.699 & 0.790 & 0.721 \\
RoBERTa & \textbf{0.924} & \textbf{0.743} & 0.625 & \textbf{0.786} & 0.789 & \textbf{0.823} \\
PHIlter & 0.854 & 0.681 & 0.216 & 0.725 & 0.728 & 0.777 \\
PyDeID & 0.788 & 0.691 & 0.314 & 0.720 & 0.698 & 0.805 \\
TiDE & 0.757 & 0.600 & - & 0.388 & \textbf{0.808} & - \\
PhysioNet-DeID & 0.704 & 0.647 & \textbf{0.811} & 0.621 & 0.798 & 0.582 \\
\bottomrule
\end{tabular}
\end{table*}

The PyDeID De-Identification algorithm had the highest recall on only name PHI for most datasets. The Transformer-DeID RoBERTa model outperformed PyDeID on the i2b2 2006 dataset and the PhysioNet De-Identification algorithm tied with PyDeID on the PhysioNet dataset. 

\begin{table*}[ht!]
\centering
\caption{Recall for each deidentification algorithm over each dataset for name labels only.}
\label{tab:recall_name}
\begin{tabular}{lrrrrrr}
\toprule
 & i2b2 2014 & i2b2 2006 & PhysioNet v2 & OpenDeID & RR-SK & RR-HHS \\
\midrule
DistilBERT & 0.689 & 0.471 & 0.303 & 0.620 & 0.157 & 0.451 \\
RoBERTa & 0.722 & \textbf{0.625} & 0.386 & 0.817 & 0.142 & 0.631 \\
PHIlter & 0.902 & 0.478 & 0.675 & 0.765 & 0.558 & 0.742 \\
PyDeID & \textbf{0.940} & 0.472 & \textbf{0.969} & \textbf{0.856} & \textbf{0.565} & \textbf{0.933} \\
TiDE & 0.702 & 0.343 & - & 0.324 & 0.539 & - \\
PhysioNet-DeID & 0.812 & 0.388 & \textbf{0.969} & 0.700 & 0.496 & 0.679 \\
\bottomrule
\end{tabular}
\end{table*}

\begin{table*}[ht!]
\centering
\caption{Number of false negatives for name PHI types per 1000 tokens.}
\label{tab:fns_name}
\begin{tabular}{lrrrrrr}
\toprule
 &     i2b2 2014 & i2b2 2006 & PhysioNet v2 & OpenDeID & RR-SK & RR-HHS \\
\midrule
DistilBERT & 3.97 & 8.37 & 1.54 & 5.74 & 4.31 & 34.89 \\
RoBERTa & 3.55 & \textbf{5.94} & 2.83 & 2.77 & 4.39 & 23.48 \\
Philter & 1.25 & 8.27 & 0.72 & 3.54 & 2.26 & 16.42 \\
PyDeID & \textbf{0.77} & 8.36 & \textbf{0.07} & \textbf{2.17} & \textbf{2.23} & \textbf{4.29} \\
TiDE & 3.80 & 10.40 & - & 10.20 & 2.36 & - \\
PhysioNet-DeID & 2.40 & 9.69 & \textbf{0.07} & 4.53 & 2.58 & 20.42 \\
\bottomrule
\end{tabular}
\end{table*}

\begin{table*}[h!]
\centering
\caption{F1-score and recall for combined RoBERTa and PyDeID algorithms using a recall-maximizing merge strategy. }
\label{tab:pydeid-roberta}
\begin{tabular}{lrrr}\toprule
&F1-score &recall \\\cmidrule{2-3}
i2b2 2014 &0.820 &0.988 \\
i2b2 2006 &0.751 &0.910 \\
PhysioNet-DeID &0.310 &0.983 \\
OpenDeID &0.770 &0.889 \\
RR-SK &0.702 &0.884 \\
RR-HHS &0.867 &0.919 \\
\bottomrule
\end{tabular}
\end{table*}

We also report the number of false negatives per 1000 tokens for name-entities (Table \ref{tab:fns_name}), which is approximately inversely proportional to recall (Table \ref{tab:recall_name}).

Next, we report the F1-score and recall when combining a machine learning- (RoBERTa) and rule-based (PyDeID) tool in Table \ref{tab:pydeid-roberta}. These annotations were merged to maximize the recall (and thus protection of patient privacy) for all datasets; i.e., if either algorithm marked a token as PHI, the merged set also labelled that token as PHI.

\section{Discussion}

Enquiries into public views on the use of patient data for research broadly suggest that there is a willingness to share data where it is for the common good. This is even true of free-text data contained in EHRs so long as they are fully deidentified \cite{ford2020}. High performance algorithms for deidentification of free-text clinical notes is a key step toward conducting research which maximizes public benefit while minimizing the risk of harm to an individual.
Here we present a software library, \emph{pyclipse}, which enables reproducible comparison of deidentification algorithms. 
The configurable evaluation enables researchers to compare different evaluation approaches and customizable annotation filters. Moreover, the framework allows integration of private datasets, enabling rapid evaluation of these open-source deidentification tools against local clinical data.

\emph{Pyclipse} provides a consistent interface to standard deidentification benchmark datasets including i2b2-2006, i2b2-2014, and OpenDeID.
Although these benchmark datasets are exceedingly useful to the broader community, the tagged entities were synthetically generated to enable sharing of the dataset without violating regulatory statues such as HIPAA.
Creating accurate synthetic data is a challenging task in itself. 
Earlier efforts such as the i2b2 2006 dataset replaced entities with intentionally adversarial entities to challenge existing systems, e.g. first and last names were replaced with unrealistic words such as ``Bell Nexbeathefarst'' and ``Linemase D. Jesc.''
Although the later datasets adopted a more realistic surrogate PHI generation approach, the use of synthetic PHI remains a potential source of error in estimating the generalization error of models.
Conversely, The two radiology reports datasets evaluated in this work (RR-HHS and RR-SK) contain real PHI. While this may provide a better evaluation of algorithm performance, they do so within a very specific domain (radiology reports) which may not generalize to other note types. 
Ultimately evaluation on local data is necessary, which motivated the utilities in \emph{pyclipse} which support mapping data into a common format for evaluation.

Five open-source deidentification algorithms were evaluated here: PyDeid, TiDE, Philter, Physionet-DeID, and Transformer-DeID.
Overall, PydeID obtained the highest recall both when evaluated on all PHI types and on PHI labeled ``name'' on most corpi examined.
Careful attention must be paid to the variability in annotated entities when interpreting the results highlighted in \ref{tab:dset_PHI_types}. For example, the i2b2-2014 corpus contains entities for all ages, rather than just those over 89, but the PhysioNet-DeID tool only annotates ages over 89.
We evaluated the models on the name entity only to mitigate this factor, and found Pydeid still had the highest recall for most datasets.
To provide an interpretable measure of model performance, we report the number of false negatives per 1000 tokens \cite{johnson2019}. As PyDeID had the highest recall, it also has the lowest FN/1000 of 0.77 as seen in Table \ref{tab:fns_name}, indicating that for every 1000 tokens processed one name token is missed. We believe this measure will allow non-experts to better interpret the practical performance of deidentification algorithms and inform decisions balancing the risk of patient harm with the benefit of improving care.
A notable exception is the i2b2-2006 corpus. As mentioned earlier, this corpus has intentionally unrealistic entities which are particularly challenging for rule-based approaches. All models had substantially lower recall on the name category of i2b2-2006, whereas the neural network based model (RoBERTa) had reasonable performance.
Recent work has highlighted the potential disparate impact of rule-based deidentification on uncommon names which are occur in underrepresented groups \cite{xiao2023}.
When annotations from PyDeID (rule-based) and RoBERTa (neural network-based) approaches are combined (Table \ref{tab:pydeid-roberta}), recall improves relative to that achieved by PyDeID alone for all datasets, indicating improved protection for patient privacy. Our comparison highlights the complementary nature of rule-based and neural-based approaches, which remains a promising avenue for future work using clinical text \cite{stubbs2019}.

For all of the algorithms, we obtain worse results on all datasets than those reported by the original paper. 
The reasons for these differences in performance are variable and differ across datasets and algorithms.
For Philter, \cite{norgeot2020} modify the gold-standard annotations to exclude years in isolation, professions, country names, ages less than 90, and hospital abbreviations whereas this evaluation does not, potentially accounting for Philter's decreased performance. 
For PhysioNet-DeID, the gold-standard annotations used in this work are sourced from an improved set published by \cite{hartman2020customization} rather than \cite{neamatullah2008} used in the original algorithm development and evaluation.

Another potential source of discrepancy is the method of tokenization and evaluation. In this paper, we report metrics calculated at the token level rather than the entity level. Entity-based evaluation requires that the predicted annotation includes the entirety of a PHI tag whereas token-level evaluation is only concerned that all parts are identified rather than being identified together \citep{stubbs2015automated}. Which method to use depends on the goals of the annotator. Entity-based evaluation is standard for named-entity recognition where capturing an entire entity is prioritized. However, for the purposes of deidentification, whether the tokens of a PHI string are labelled together is less relevant for protecting patient privacy than all parts being labelled. Similarly, the tokenizer itself can have significant impacts on evaluation performance, which has historically been underappreciated.

A major goal of the construction of deidentification algorithms is to be able to apply them across institutions, specialties, and contexts. 
The results presented here indicate that much work remains to enable this sort of algorithmic generalization. 
Take, for instance, the model and dataset with the highest F1-score performance: the RoBERTa model on the i2b2 2014 dataset, which achieved an F1-score of $0.92$. 
On all other datasets, the RoBERTa model’s F1-score did not exceed 0.8, making it clearly currently insufficient for generalization. In particular, RoBERTa (and the other Transformer-DeID models) performed extremely poorly on name-only recall for both the PhysioNet and RR-SK corpi, potentially because in the i2b2 2014 training dataset, most names occur at the top of the document near other PHI such as medical record number, whereas in the RR-SK and PhysioNet datasets, names appear in a less structured and less consistent manner. 

To demonstrate the utility of \emph{pyclipse}, we evaluated the algorithms on two privately held datasets: RR-SK and RR-HHS (Table \ref{tab:f1}). The two algorithms with the highest F1-score on the RR-SK dataset, PhysioNet and DistilBERT (excluding TiDE for comparison), achieved the lowest F1-score on the RR-HHS dataset.
This variation in performance highlights the need for local evaluation datasets targeted at the text and clinical domain of interest. Interestingly, this discrepancy is not strong in the comparison of name-only recall for the datasets in Tables \ref{tab:recall_name} and \ref{tab:fns_name}, indicating that generalizability may differ between entity types.

This work has a number of important limitations. 
Firstly, we chose not to evaluate deidentification algorithms which explicitly require customization and fine-tuning using a local dataset such as MIST \cite{aberdeen2010}.
Our goal with pyclipse was to provide a rapid interface to compare existing deidentification approaches, and the custom nature of these approaches prohibit a general comparison.
We further elected to only include algorithms which were open source, precluding a number of top performing algorithms submitted to the i2b2 2014 challenge \cite{stubbs2015automated, lee2017}, as well as proprietary services offered by cloud providers.
Secondly, we were limited in the clinical text available for evaluation due both to usage restrictions and the paucity of existing annotated text. 
Though we have made an effort to include a range of note types (discharge summaries, nursing notes, pathology reports, and radiology reports), deidentification research is still limited by the dearth of annotated text data containing real or surrogate PHI.
For instance, the 2016 N-GRID challenge found decreased performance in many established deidentification approaches due to the data shift toward psychiatric records \cite{stubbs2017}.
The local data evaluation only included notes from a specific clinical domain (radiology) had a low number of samples relative to publicly available datasets, reducing precision in the performance estimates.
Including broader note types such as psychiatric records would enhance the inferences possible from \emph{pyclipse}.

Our study has a number of strengths.
We have containerized a number of open-source deidentification approaches, enabling rapid deployment with minimal setup required.
Our transparent evaluation pipeline allows a consistent assessment of the different approaches, and we have provided an extensive comparison of five open-source algorithms across six datasets.
Finally, our modular design allows extension of \emph{pyclipse} to additional algorithms and local datasets.

\section{Conclusion}

A major impediment to deidentification research is the tension between protecting patient privacy by controlling access to sensitive patient data while enabling shared discovery across the research community.
Given these constraints, we created \emph{pyclipse}, a Python library which enables uniform evaluation of multiple algorithms over public and local datasets.
We found algorithm performance was consistently lower than the results reported by the original papers, even on the same benchmark dataset. These differences may be attributed to variability in evaluation protocols, annotation schemes, and technical implementation.
Our work highlights the complexity and challenge of accurately assessing and comparing deidentification algorithms, hence demonstrating the necessity and utility of a reproducible, adjustable and extensible framework such as \emph{pyclipse}.
Our work provides a stepping stone towards a unified approach in evaluating and refining deidentification tools, facilitating the protection of patient data.

\bibliographystyle{plain} 
\bibliography{reference} 

\end{document}